\documentclass[10pt,twocolumn,letterpaper]{article}
\usepackage{iccv}
\usepackage{times}
\usepackage{epsfig}
\usepackage{graphicx}
\usepackage{amsmath}
\usepackage{amssymb}
\usepackage{flushend}
\usepackage{authblk}
\usepackage{setspace}
\usepackage{subfig}
\usepackage{caption}

\usepackage{hyperref}

\iccvfinalcopy 

\def\iccvPaperID{13}

\ificcvfinal\fi
\begin{document}

\title{What are the visual features underlying human versus machine vision?}


\author{D. Linsley* \qquad S. Eberhardt \qquad T. Sharma \qquad P. Gupta \qquad T. Serre \\
Brown University\\
Providence, RI, USA\\
{\tt\small *drew\_linsley@brown.edu}\\
\textit{Appearing in the proceedings of the Mutual Benefits of Cognitive and Computer Vision (ICCVw 2017).}
}
\maketitle

\begin{abstract}
    Although Deep Convolutional Networks (DCNs) are approaching the accuracy of human observers at object recognition, it is unknown whether they leverage similar visual representations to achieve this performance. To address this, we introduce Clicktionary, a web-based game for identifying visual features used by human observers during object recognition. Importance maps derived from the game are consistent across participants and uncorrelated with image saliency measures. These results suggest that Clicktionary identifies image regions that are meaningful and diagnostic for object recognition but different than those driving eye movements. Surprisingly, Clicktionary importance maps are only weakly correlated with relevance maps derived from DCNs trained for object recognition. Our study demonstrates that the narrowing gap between the object recognition accuracy of human observers and DCNs obscures distinct visual strategies used by each to achieve this performance.
\end{abstract}

\section{Introduction}


Advances in Deep Convolutional Networks (DCNs) have led to vision systems that are starting to rival human accuracy in basic object recognition tasks \cite{He2016-oe}. While a growing body of work suggests that this surge in performance carries concomitant improvement in fitting both neural data in higher areas of the primate visual cortex (reviewed in \cite{Yamins2016-yx}) and human psychophysical data during object recognition \cite{Kheradpisheh2016-dm,Peterson2016-ig}, key differences remain. 

It has been suggested that the processing depth achieved by state-of-the-art DCNs may be greater than that achieved by the human visual system during rapid categorization \cite{Eberhardt2016-hc}. It has also been shown that DCNs do not generalize well to atypical scenes, such as when objects are presented outside of their usual context \cite{Saleh2016-hs}. A recent study \cite{Ullman2016-ea} found evidence for qualitatively different patterns of behavior by human observers versus DCNs during recognition. When presented with small object crops, human participants depended on the inclusion of a key diagnostic image feature to recognize the objects. In contrast, DCNs failed to exhibit the same ``all-or-nothing'' dependence on key visual features during object recognition. Overall, this raises the possibility that DCNs may leverage entirely different visual strategies than humans during object recognition. 

\begin{figure}[t!]
\begin{center} 
   \includegraphics[width=1\linewidth]{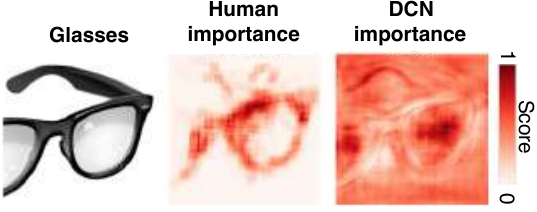}
\end{center}
   \caption{Humans and DCNs utilize different visual features for object recognition. We developed an online game called \emph{Clicktionary} to derive human importance maps for object recognition, which we compared to feature importance maps derived from DCNs.}
 \label{example_importance:1}
\end{figure}


Here, we provide direct evidence that the visual features used by DCNs for object recognition differ markedly from those used by human observers. We created \emph{Clicktionary}, a collaborative web-based game for identifying diagnostic visual features for human object recognition. Pairs of participants work together to identify objects: One player reveals diagnostic image regions while the other tries to recognize the object as quickly as possible from those image parts. Amassing game-play data across many participants yields importance maps for individual images. This is illustrated in the middle panel of Figure~\ref{example_importance:1}: the hotter the pixel, the more often it was selected by participants as important for recognition. For this image of glasses, the frame was more important for recognition than its lenses. Thus, these feature importance ''hot spots'' highlight key features human observers use to recognize object images.

Using importance maps derived from \emph{Clicktionary} we show that: {\bf (1)} Features identified in these maps are strongly stereotyped across participants and also different from those in saliency maps derived from both attention models and human participants. {\bf (2)} Humans and DCNs favor dissimilar visual features during object recognition, revealing a novel difference between biological and machine vision and an opportunity to bridge this gap.

\section{Related Work} 

\paragraph{Behavioral studies:} A central goal in vision science is to understand what features the human visual system uses to process complex visual scenes. Traditional approaches to uncover the internal representations contributing to behavior include reverse correlation methods, which involves analyzing the relationship between decisions made about a stimulus and visual perturbations applied to it across many trials. Reverse correlation methods have helped identify visual features that are diagnostic for faces and other synthetic object stimuli \cite{Schyns2002-xt, Nielsen2008-ey}. But these methods are inefficient and typically require thousands of trials per subject to deduce internal representations despite little shape variability in these object classes. Reverse correlation approaches therefore seem impractical for characterizing visual representations used for categorizing general classes of objects with higher variability in stimulus appearance, location, or lighting.

Recording eye fixations while viewing a stimulus is another way of exploring visual feature importance. Patterns of eye fixations represent observers' efforts to center their high-acuity fovea on salient or task-relevant information in images \cite{Yarbus1967-xp}. Eye fixations are typically recorded during passive image viewing, making it difficult to associate them with object recognition. It is also difficult and costly to acquire large-scale eye tracking data, leading researchers instead to track computer mouse movements during task-free viewing of images to estimate local saliency cues for fixations \cite{Lin_undated-qc, Jiang2015-dr}. 


Cognitive psychologists have traditionally used similarity judgments between image pairs to study visual representations. Recent work has used this approach to compare representations between humans and representative DCNs, finding good agreement between the two \cite{Peterson2016-ig}. Related work has also evaluated the ability of DCNs to predict memorability \cite{Dubey2015-wy} and the typicality of individual images \cite{Lake2015-jw}.



 \paragraph{Computational models:} A growing body of research sought understanding of the visual features used by DCNs for object recognition. There is no gold standard, but popular methods fall into one of three groups. Sensitivity analyses use either gradient-based approaches \cite{Zeiler2013-mk, Simonyan2014-ap} or systematic perturbations of the stimulus to estimate local pixel-wise contributions of visual features to a classification decision \cite{Zeiler2013-mk}. Decision analyses such as layer-wise relevance propagation (LRP) provide a global estimate of a pixel's responsibility to the classification decision. A third approach adopted by methods like class activation mapping (CAM) is to optimize DCN visualizations for class-discriminability, yielding reliable object localization \cite{zhou2015cnnlocalization}. A representative methods from each approach is used here to derive importance maps from DCNs for comparison with those derived from human observers.

 \paragraph{Web-based games for data collection:} There is a history of leveraging the wisdom of crowds through web-based applications to gather data for computer-vision studies. Closely related to \emph{Clicktionary} is the ESP game for identifying objects in real-world images \cite{Von_Ahn2004-qe} and the Peek-a-boom game for locating them \cite{Von_Ahn2006-bz}. In both of these games, participants work together to recognize an image of an object. We take particular inspiration from Peek-a-boom, where one participant in a pair reveals parts of an image to elicit a classification response from the other participant.

\emph{Clicktionary} alters the mechanics of Peek-a-boom in two keys ways that make it more suitable for measuring feature importance in object images. {\bf (1)} \emph{Clicktionary} has a high resolution interface that lets participants more selectively reveal visual features for object recognition. {\bf (2)} \emph{Clicktionary} controls for image revelation strategies from teachers that could introduce confounds into the resulting feature importance maps, such as ``salt and peppering" the screen with clicks, waiting long intervals between clicks, or sending visual hints back and forth as in the original Peek-a-boom game.

Another web-based game with a similar goal as \emph{Clicktionary} is Bubbles \cite{Deng2013-pc}, which surveys features in images useful for distinguishing between two object categories. In this single-player game, participants first familiarize themselves with multiple ``training'' exemplars from two image categories. Participants are then presented with a blurred ``test'' exemplar and asked to sharpen pixels that are most informative for identifying the correct category. This design does not effectively scale to the number of object categories that we test here, and the interface obscures the extent to which revealed local features versus the ``gist'' of the blurred test image supported its recognition.

Similar game-like interfaces have also been used to explore other facets of human perception as in \cite{Steggink2010-lb} where players take turns outlining important objects in real-world scenes and guessing their identities. Another example is \cite{Das2016-da}, where participants sharpened parts of a scene image that they deemed important for answering questions about the scene. The game style has also been used to answer questions in biological and physical sciences, including predicting protein structure \cite{Cooper2010-jq} or neuronal connectivity \cite{Kim2014-fp}.

\begin{figure}[t]
\begin{center}
   \includegraphics[width=1\linewidth]{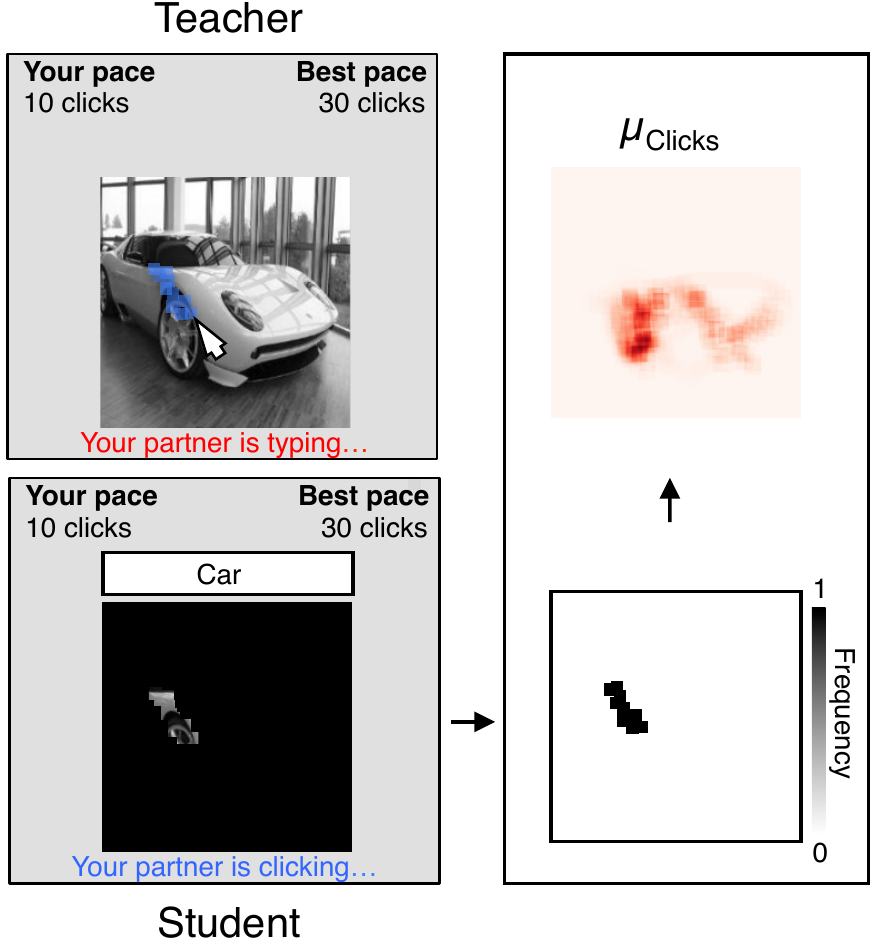}
\end{center}
   \caption{Overview of \emph{Clicktionary}. Pairs of participants, one teacher and one student, play together to categorize objects. The teacher uses the mouse to place ``bubbles'' on the image to reveal regions to the student, who types the category name of the object. Bubble densities computed across participants at each pixel location yields importance maps that identify visual features diagnostic for recognition.}
 \label{clicktionary_overview}
\end{figure}

\section{The Clicktionary game}

\emph{Clicktionary} was constructed to identify visual features that are diagnostic for human observers during visual recognition. Upon starting the game, players provided informed consent, read instructions, and were placed into a virtual waiting room. The waiting room contained a scoreboard listing the performance of the most successful teams to play the game. Our hope was to provide an incentive for players to compete with each other in order to collect the highest quality behavioral data possible. Participants were automatically paired in the waiting room\footnote{If no one else entered the waiting room within 120 secs, players played against a DCN opponent. These data were not included in the current study.}. Once the game began, players collaborated over a series of rounds to name the category of an object image (Figure~\ref{clicktionary_overview}). 

Players accomplished this by alternating between two different roles over many game rounds. In each round, one player was the teacher, who viewed the intact image, and the other was the student, who viewed a blank image. The teacher revealed regions of the object in the image that were thought to be informative for the student to recognize it (whose role is detailed below).

The teacher revealed informative image regions by clicking on the image and then dragging the mouse across it, as if painting. Translucent blue boxes on the teacher's image marked regions that were visible to the student. Each of these image regions, revealed to the student by the teacher, were $18\times18$ pixels. We refer to these revealed image patches hereafter as ``bubbles" \cite{Gosselin2001-te,Deng2013-pc}. Images were $300\times300$ pixels.

The mechanics controlling how teachers bubbled images were constructed to maximize our ability to identify minimal object features. Teachers were instructed to carefully choose where they began bubbling because once they started, the process did not stop until the student recognized the image. Bubbles were continuously placed at the position of the teacher's mouse cursor, leading to the feeling of ``painting'' these bubbles on the image. To ensure that all teachers bubbled at a relatively consistent rate, bubbles appeared under the mouse cursor at a random time interval ranging between 50 msecs and 300 msecs. 

Each bubble after the first was placed within the radius of the one preceding it. This meant that teachers could easily and precisely bubble in the image, but could not change the speed at which the bubbles were placed. This also kept teachers from using a ``salt-and-pepper'' strategy to skip around the screen when bubbling. We felt this design choice was crucial for controlling against this and other bubbling strategies that would likely have yielded faster game play but also hindered our ability to capture minimal object features. For example, it is possible that students could have inferred object category clues from the shapes created by the bubbles themselves.

In contrast to the teacher, the student began each round viewing a blank version of the teacher's image and a text box for guessing its object category. As the teacher bubbled image regions, corresponding locations of the student's blank image were unveiled to reveal content. Note that having students begin each round with a blank rather than a blurred version of the image ensured that they only derived object information from image regions bubbled by their teachers. Students were instructed to name the basic-level category of the object. For instance, the desired response for an image of a border collie was ``dog". We also accepted subordinate-level category labels to expedite game play. 

To provide teams of players incentive to work as quickly and efficiently as possible, their performance and the average performance of the top-10 scoring teams was visible throughout the game. Team performance was measured as the number of image bubbles placed by teachers before students recognized the image. While there was no explicit penalty for wrong answers, participants achieved better scores by avoiding them. If a team finished the game in the top-10 they were congratulated and shown their ranking.

\begin{figure*}[t]
\begin{center}
   \includegraphics[width=1\linewidth]{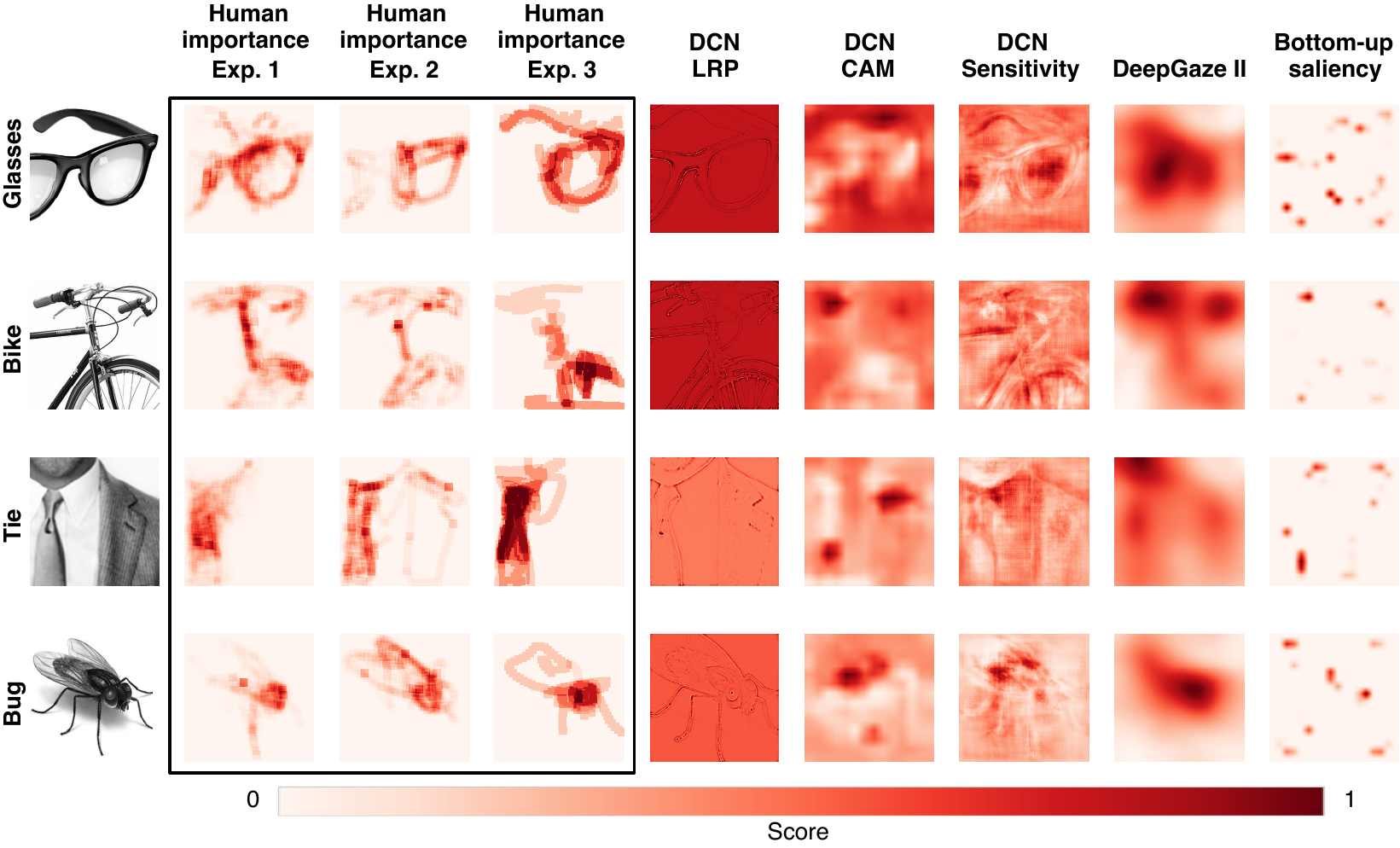}
\end{center}
   \caption{Heatmaps depicting object feature importance for humans and machines. From left to right: importance maps on images from Experiments 1, 2, and 3 (the black box was added for emphasis), layer-wise relevance propagation maps (LRP) from VGG16, class activation mapping (CAM) on VGG16, sensitivity analysis on VGG16, predicted saliency from DeepGaze~II, and bottom-up saliency.}
 \label{mirc_heatmaps}
\end{figure*}

Incorrect guesses caused a red outline to appear around the image viewed by both student and teacher. For correct guesses, images were briefly outlined in green before proceeding to the next round. If a student could not figure out the object's class, there was a skip button that penalized the team's performance with the equivalent of 100 bubbles. Student and teacher switched roles after each round. The game was played for 110 rounds, with a different image each round. Each pair played on a random ordering of images. 

By design participants could not communicate, but we included features to make the game feel more collaborative. These included real-time notifications of what each player in the pair was doing at any point in time: bubbling, typing, correct and incorrect responses, or considering which part of the image to bubble first. 

Each game lasted about 20 minutes and participants were only allowed to play once. Participants were recruited through Amazon Mechanical Turk or from introductory computer science or cognitive science classes, and reimbursed approximately \$8.00/hr.




We created importance maps through a two-step procedure. First, individual bubble maps were created for every image played by a \emph{Clicktionary} participant pair (lower-right corner, Figure~\ref{clicktionary_overview}). Second, feature importance maps were derived as the average number of bubbles at each image pixel across all participant pairs (top-right corner, Figure~\ref{clicktionary_overview}).

\section{Comparing importance maps derived from humans versus DCNs}

We validated the reliability of feature importance maps derived from \emph{Clicktionary} and used them to systematically compare the features used by humans and DCNs to categorize representative objects. We used rank-order correlation throughout to measure these associations because importance maps are sparse and do not satisfy normality assumptions (no click map passed a Kolmogorov-Smirnov test for normality). All tests for significance used independent sample \textit{t}-tests with two-tailed \textit{p}-values.

\paragraph{Reliability of Clicktionary data:}

We first ran two rounds of the \emph{Clicktionary} game which are hereafter referred to as Exp. 1 and Exp. 2. Forty-six participants took part in Experiment 1 and 14 participants in Experiment 2. Both Exp. 1 and Exp. 2 included the ten original images used in \cite{Ullman2016-ea}, for possible comparison with the MIRCs (graciously provided by the authors; data not shown). In addition, each experiment had participants judge a different set of 100 object images taken from the validation set of the 2012 ImageNet Large Scale Visual Recognition Challenge (ILSVRC) \cite{Russakovsky2014-ff}. Images for each experiment were selected from 10 categories, 5 animate and 5 inanimate. For Exp. 1, we chose representative categories for animate objects: border collie, bald eagle, great white shark, and sorrel; and inanimate objects: airliner, school bus, speedboat, sports car, and trailer truck. For Exp. 2, we chose the 5 animate and 5 inanimate categories that were the most difficult for VGG16 \cite{Simonyan2015-pj} to categorize (top-1 accuracy). These were english foxhound, husky, miniature poodle, night snake, and polecat; cassette player, missile, screen, sunglasses, and water jug. 


We measured the inter-participant and inter-experiment consistency of the importance maps extracted from the \emph{Clicktionary} game to validate the effectiveness of these maps at capturing visual features for object recognition. Inter-participant consistency was measured by splitting participants into two random groups and then recording the rank-order correlation between each group's mean importance maps. Each experiment's inter-participant consistency was found by taking the mean score across 1000 iterations of this procedure. We found $\rho=0.88$ (\textit{p} \textless 0.001) in Exp. 1 and $\rho=0.79$ (\textit{p} \textless 0.001) in Exp. 2. We also measured consistency between participants in Exp. 1 \vs Exp. 2 on the MIRC images that each group saw. Again, there was a strong correspondence between importance maps derived from these 10 images for participants across the two independent experiments even though Exp. 2 had roughly a third as many participants as Exp. 1 ($\rho=0.55$, \textit{p} \textless 0.001).

\begin{figure}[t]
\begin{center}
   \includegraphics[width=1\linewidth]{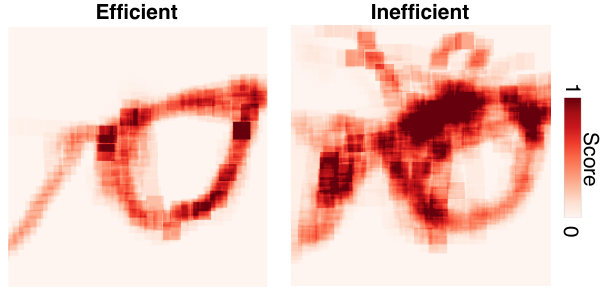}
\end{center}
   \caption{Importance map features vary according to how efficiently a participant pair recognized images. On the left, the mean importance map from pairs with above-median efficiency in recognizing glasses (\ie faster recognition). On the right, the mean importance map from below-median pairs. The above image is representative of the typical differences found between these groups.}
\vspace{0mm}\label{split_half_viz}
\end{figure}

Despite the strong agreement we observed between \emph{Clicktionary} participants, we found that importance maps were affected by player performance. Applying a median split to the number of bubbles it took for pairs to recognize objects revealed qualitatively different importance maps for the two groups (Figure~\ref{split_half_viz}). As expected, maps from the efficient group (\ie number of bubbles below the median split) had significantly stronger inter-participant correlations than the inefficient group for both experiments (Exp. 1: $\rho_{\text{EFFICIENT}}$ = 0.94 versus $\rho_{\text{INEFFICIENT}}$ = 0.89, \textit{p} \textless 0.001, and Exp. 2: $\rho_{\text{EFFICIENT}}$ = 0.92 versus $\rho_{\text{INEFFICIENT}}$ = 0.84, \textit{p} \textless 0.001). The efficient group also yielded quantitatively sparser maps than the inefficient group, which we measured as the kurtosis of each groups mean importance map for every image\footnote{Because kurtosis measures distributional characteristics, it is robust to the total number of clicks made by each group and controls for potential circularity in this analysis.}. For both experiments, the average kurtosis across images was significantly greater for the efficient group than the inefficient group, indicating that the distribution of the efficient group's importance maps were more peaked and carried stronger ``hotspots'' than the inefficient group: Exp. 1: $\beta_2{_{\text{EFFICIENT}}}$ = 49.51 versus $\beta_2{_{\text{INEFFICIENT}}}$ = 34.47, \textit{p} \textless 0.001; Exp. 2: $\beta_2{_{\text{EFFICIENT}}}$ = 20.27 versus $\beta_2{_{\text{INEFFICIENT}}}$ = 9.73, \textit{p} \textless 0.001). These results suggest that although \emph{Clicktionary's} feature importance maps are stereotyped across participants, the underlying visual strategies are nevertheless somewhat varied. More work is needed to better characterize the exact of visual features and strategies used to generate these maps.

\begin{figure*}[t]
\begin{center}
   \includegraphics[width=.75\linewidth]{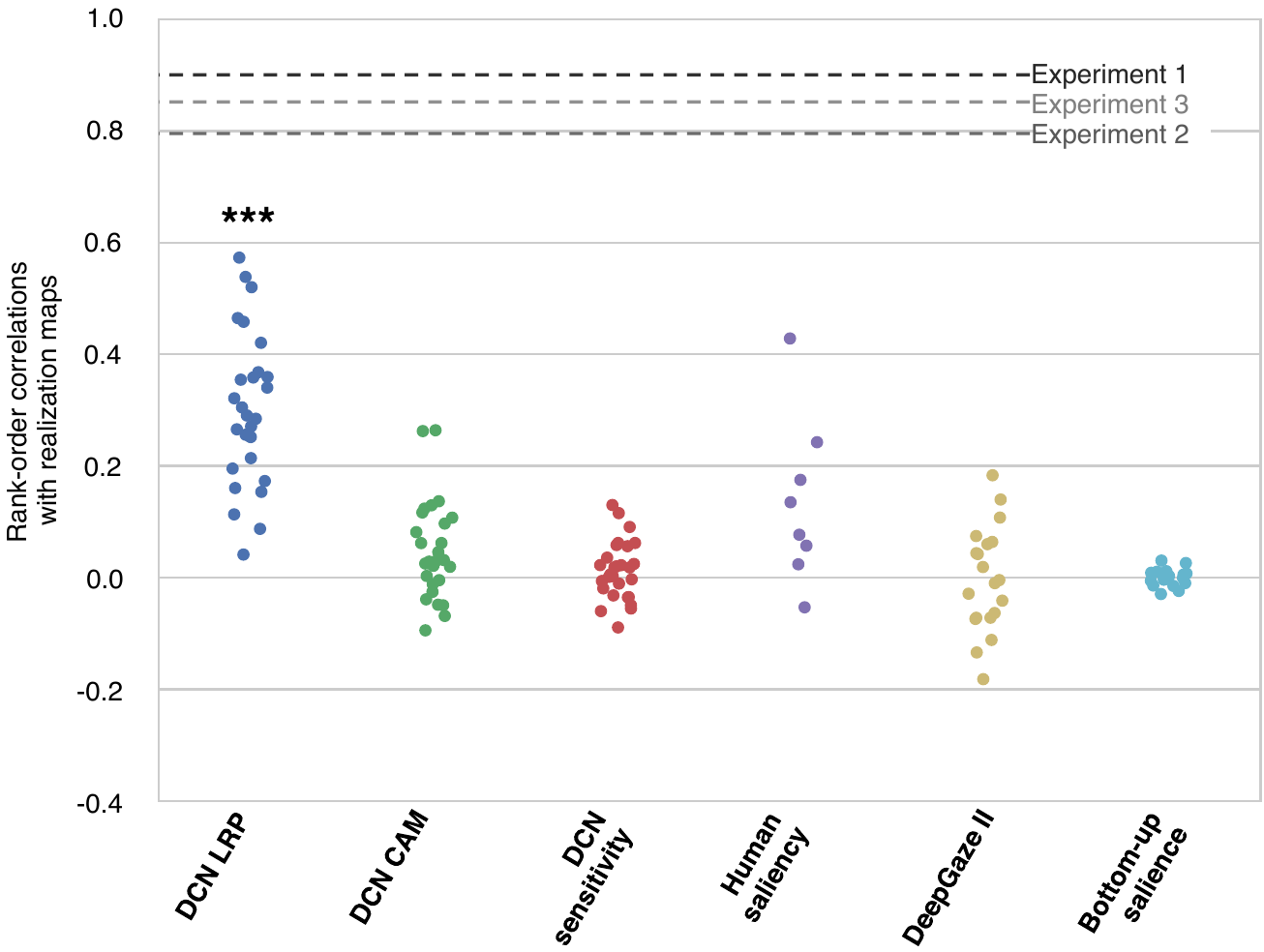}
\end{center}
   \caption{Correlations between  \emph{Clicktionary} feature importance maps, DCN feature importance, and visual saliency. Each dot represents the mean correlation between \emph{Clicktionary} and a feature importance source for individual image categories. For reference, inter-participant correlations are plotted in dashed lines for Experiment 1, Experiment 2, and Experiment 3. Not all image categories are represented by each feature importance source: importance maps from a bottom-up saliency algorithm and DeepGaze II were derived for images from Experiments 1 and 2, whereas human saliency was only measured on images in Experiment 3. Independent samples \textit{t}-tests measured deviation from 0. ***: \textit{p} \textless 0.001} 
 \label{imagenet_corr_scatters}
\end{figure*}

 \paragraph{Weak correlation between Clicktionary and DCN data:} A key objective of this study was to compare importance maps derived from human participants and DCNs. In support of this, we produced DCN heatmaps of object importance for each of the images used in \emph{Clicktionary}. This was done using VGG16, a variant of the popular VGG architecture \cite{Simonyan2015-pj}. We calculated heatmaps for this model using three representative methods: a sensitivity analysis \cite{Zeiler2013-mk}, LRP \cite{Bach2015-dg}, and CAM \cite{zhou2015cnnlocalization}. 

Strikingly, there was only a weak relationship between \emph{Clicktionary} importance maps and LRP derived from VGG16 ($\rho = 0.33$, \textit{p} \textless 0.001; Figure~\ref{imagenet_corr_scatters}). Although this correlation was significantly greater than 0, it was significantly weaker than each experiment's inter-participant correlations. This was measured as the proportion of iterations of the inter-participant randomization procedure described above that yielded smaller correlations than the mean LRP correlation: inter-participant Exp. 1 $>$ LRP, \textit{p} \textless 0.001; inter-participant Exp. 2 $>$ LRP, \textit{p} \textless 0.001. In addition, importance maps did not correlate with feature importance maps from a sensitivity analysis performed on VGG16 ($\rho = 0.03$, n.s.; inter-participant Exp. 1 $>$ sensitivity, \textit{p} \textless 0.001; inter-participant Exp. 2 $>$ sensitivity, \textit{p} \textless 0.001). 

We also produced DCN heatmaps based on the CAM method \cite{zhou2015cnnlocalization} because it is designed to localize objects rather than reflect DCN decision processes. Implementing CAM requires modifying a DCN, replacing its fully connected layers with a single layer that maps its convolutional features to categories. We trained the CAM layer for 100 epochs on a subset of ILSVRC12 (150,000 images total), and selected the weights that yielded maps with the strongest correlation with human feature importance maps derived from \emph{Clicktionary}. Even with this optimization procedure in place, feature importance maps from CAM were uncorrelated with either \emph{Clicktionary} experiment. We found a correlation $\rho=0.10$ (n.s.) in Exp. 1 and $\rho=0.06$ (n.s.) in Exp. 2 (inter-participant Exp. 1 $>$ CAM, \textit{p} \textless 0.001; inter-participant Exp. 2 $>$ CAM, $p < 0.001$).

\paragraph{Lack of association between Clicktionary data and saliency measures:} 

We considered the extent to which \emph{Clicktionary} maps were consistent with visual saliency maps. Because theory holds that attention is driven by both bottom-up and top-down mechanisms \cite{Itti2001-kw}, we compared importance maps to saliency maps derived from models for both kinds of attention, referred to hereafter as bottom-up saliency \cite{Itti2001-kw}\footnote{The algorithm for predicting bottom-up saliency maps was tuned to have qualitatively similar sparsity as the importance maps.} and top-down attention derived from the DeepGaze~II model \cite{Kummerer2016-vp}. 

Importance maps were not correlated with either bottom-up saliency ($\rho = 0.00$, n.s.) or eye fixation predicted by DeepGaze~II ($\rho = 0.03$, n.s.; Figure~\ref{imagenet_corr_scatters}). Inter-participant correlations for each experiment were also significantly stronger than correlations between \emph{Clicktionary} feature importance maps and either measure of saliency (inter-participant Exp. 1 $>$ bottom-up saliency, \textit{p} \textless 0.001; inter-participant Exp. 2 $>$ bottom-up saliency, \textit{p} \textless 0.001; inter-participant Exp. 1 $>$ DeepGaze~II, \textit{p} \textless 0.001; inter-participant Exp. 2 $>$ DeepGaze~II, \textit{p} \textless 0.001).

To better estimate the similarity between \emph{Clicktionary} feature importance maps and visual saliency we ran an additional round of the \emph{Clicktionary} game (hereafter referred to as Exp. 3)
to directly compare its importance maps to human saliency maps from the SALICON dataset \cite{Jiang2015-dr}. SALICON contains a subset of images from the Microsoft COCO image dataset, and for exp. 3 we used images from categories that were also in ILSVRC, which allowed us to also compare feature importance maps from this experiment with DCNs. In total Exp. 3 had 8 total object categories, 4 animate and 4 inanimate, as well as the 10 images from \cite{Ullman2016-ea} (90 total). Image categories were bird, cat, elephant, and zebra; couch, dining table, refrigerator, and umbrella.

\begin{figure}[t]
\begin{center}
   \includegraphics[width=1\linewidth]{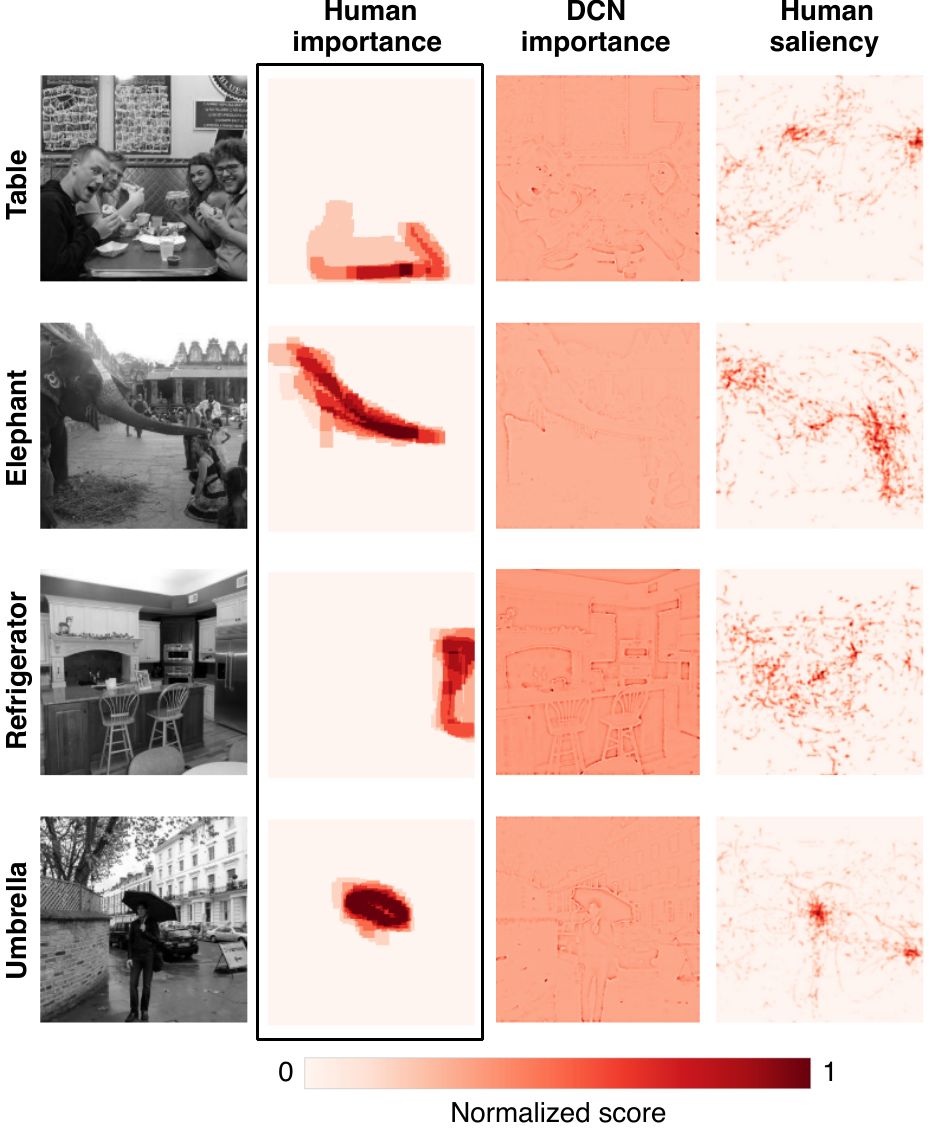}
\end{center}
   \caption{Human feature importance maps from Exp. 3 contrasted with human saliency maps and DCN importance (LRP) for the same images. Note that human feature importance maps from \emph{Clicktionary} are both qualitatively unique from the others and strongly stereotyped.}
\vspace{0mm}\label{fig_mirc_corr}
\end{figure}

As with the other two \emph{Clicktionary} experiments, there was strong inter-participant correlation for the 12 participants in Exp. 3 ($\rho=0.85$, \textit{p} \textless 0.001). Consistent with the computational models for visual saliency, there was weak correlation between \emph{Clicktionary} and human saliency data: $\rho=0.14$ (\textit{p} \textless 0.001; even after adjusting the saliency maps to have similar resolution as the importance maps; Figure \ref{fig_mirc_corr}).



\section{Discussion}
\emph{Clicktionary} is a novel approach for estimating feature importance maps derived from human participants. The proposed method overcomes some of the limitations of existing psychophysical methods including reverse correlation and other image classification methods for measuring internal representations used by human observers to recognize objects (see \cite{Murray2011-hl} for a review). We have described what is, to our knowledge, the first systematic study of feature importance maps derived from human participants using natural images over multiple object categories.


To summarize our main findings: We found little or no overlap between importance maps and image saliency maps predicted by attention models (Exp. 1 and 2) or derived from human participants (Exp. 3). This suggests that importance maps reflect neural mechanisms that are, to some degree distinct from those that guide attention and eye movements. This is a necessary distinction to make because it means that importance maps can provide insight into our understanding of biological vision in ways that extant computational models cannot.

It is important to note that importance maps are generated through a process that necessarily depends on attention: Teachers' clicks reflect a continuous ranking of the importance of image features for recognition. However, this is distinct from typical methods for measuring (or predicting) attention in two ways. First, saliency is usually measured passively or in a search task, whereas importance maps take advantage of the teacher's ability to select relevant features, trading image saliency for feature diagnosticity. Second, the interplay of teacher and student identifies the point at which visible features become sufficient to trigger recognition.

Our current method for visualizing importance maps across participant pairs remains relatively simple -- potentially disregarding important information through averaging. We expect that future work in developing a more nuanced approach to measuring importance maps and better characterizing students' recognition processes will prove useful for further characterizing visual strategies for both humans and DCNs. In support of this, \emph{Clicktionary} feature importance maps can be downloaded from \url{clickme.ai/about}.

We found a weak association between importance maps and LRP maps derived from VGG16. This indicates that there is at least some overlap between the visual features used by humans and DCNs for object categorization. However, the magnitude of this association was less than half of what was found within \emph{Clicktionary} participants, demonstrating that object representations of DCNs and humans are still meaningfully different. 

These findings support our key assertion: visual strategies used by humans and DCNs during object recognition are not aligned. Importance maps capture mostly distinct information from DCNs, as measured by either LRP or a sensitivity analysis. \emph{Clicktionary} feature importance maps were also at best weakly associated with either model- or human-derived salience maps of object images. 


How can we close the gap between human and DCN vision? One possibility is by creating a dataset of importance maps that is large and diverse enough to be included in object recognition training routines of DCNs. We have created \url{clickme.ai} to achieve this goal, and future work incorporating its feature maps into object recognition training routines represents a novel opportunity to rectify the mismatch in visual strategies between humans and DCNs revealed by \emph{Clicktionary}.

There is strong evidence that human-derived information can help machine performance in vision tasks, particularly for images depicting atypical views or containing many occlusions. Researchers have found that in cases like these human perceptual judgments can augment the performance of vision models \cite{Vondrick2015-pr}, resulting in significant gains in face recognition and localization \cite{Scheirer2014-gp, Branson2010-dr, Kovashka2016-oi}, action recognition \cite{Vig_undated-jg}, object detection and segmentation \cite{Papadopoulos_undated-tp, Shanmuga_Vadivel2015-wi, Vijayanarasimhan2014-em}.

Overall, the present work makes significant contributions to our understanding of biological vision and reveals a significant gap between feature importance for humans and machines. We believe that these findings will inspire new directions in DCN research and help narrow the gap between biological and computational vision.

\section*{Acknowledgements}
The original idea for \emph{Clicktionary} was suggested by Prof. Todd Zickler (Harvard University). We are also thankful to Matthias K{\"u}mmerer for running DeepGaze on our images and for Danny Harari for providing MIRC stimuli used in \cite{Ullman2016-ea}. We are also indebted to Junkyung Kim for comments on the manuscript. This work was supported by NSF early career award (IIS-1252951) and DARPA young faculty award (N66001-14-1-4037).

{\small
\bibliographystyle{ieee}
\bibliography{Refs}

\begin{thebibliography}{10}\itemsep=-1pt

\bibitem{Bach2015-dg}
S.~Bach, A.~Binder, G.~Montavon, F.~Klauschen, K.-R. M{\"{u}}ller, and
  W.~Samek.
\newblock On {pixel-wise} explanations for {non-linear} classifier decisions by
  {Layer-Wise} relevance propagation.
\newblock {\em PLoS One}, 10(7):e0130140, 10~July 2015.

\bibitem{Branson2010-dr}
S.~Branson, C.~Wah, F.~Schroff, B.~Babenko, P.~Welinder, P.~Perona, and
  S.~Belongie.
\newblock Visual recognition with humans in the loop.
\newblock In {\em European Conference on Computer Vision {(ECCV)}}, pages
  438--451. Springer, Berlin, Heidelberg, 5~Sept. 2010.

\bibitem{Cooper2010-jq}
S.~Cooper, F.~Khatib, A.~Treuille, J.~Barbero, J.~Lee, M.~Beenen,
  A.~Leaver-Fay, D.~Baker, Z.~Popovi\'{c}, and F.~Players.
\newblock Predicting protein structures with a multiplayer online game.
\newblock {\em Nature}, 466(7307):756--760, 5~Aug. 2010.

\bibitem{Das2016-da}
A.~Das, H.~Agrawal, C.~Lawrence~Zitnick, D.~Parikh, and D.~Batra.
\newblock Human attention in visual question answering: Do humans and deep
  networks look at the same regions?
\newblock 11~June 2016.

\bibitem{Deng2013-pc}
J.~Deng, J.~Krause, and L.~Fei-Fei.
\newblock {Fine-Grained} crowdsourcing for {Fine-Grained} recognition.
\newblock In {\em {IEEE} Conference on Computer Vision and Pattern Recognition
  {(CVPR)}}, pages 580--587, June 2013.

\bibitem{Dubey2015-wy}
R.~Dubey, J.~Peterson, A.~Khosla, M.-H. Yang, and B.~Ghanem.
\newblock What makes an object memorable?
\newblock In {\em Proceedings of the {IEEE} International Conference on
  Computer Vision}, pages 1089--1097, 2015.

\bibitem{Eberhardt2016-hc}
S.~Eberhardt, J.~Cader, and T.~Serre.
\newblock How deep is the feature analysis underlying rapid visual
  categorization?
\newblock In {\em Neural Information Processing Systems}, 2016.

\bibitem{Gosselin2001-te}
F.~Gosselin and P.~G. Schyns.
\newblock Bubbles: a technique to reveal the use of information in recognition
  tasks.
\newblock {\em Vision Res.}, 41(17):2261--2271, Aug. 2001.

\bibitem{He2016-oe}
K.~He, X.~Zhang, S.~Ren, and J.~Sun.
\newblock Deep residual learning for image recognition.
\newblock In {\em Computer Vision and Pattern Recognition}, 2016.

\bibitem{Itti2001-kw}
L.~Itti and C.~Koch.
\newblock Computational modelling of visual attention.
\newblock {\em Nature Review Neuroscience}, 2(3):194--203, 2001.

\bibitem{Jiang2015-dr}
M.~Jiang, S.~Huang, J.~Duan, and Q.~Zhao.
\newblock {SALICON}: Saliency in context.
\newblock In {\em {IEEE} Conference on Computer Vision and Pattern Recognition
  ({CVPR})}, pages 1072--1080, June 2015.

\bibitem{Kheradpisheh2016-dm}
S.~R. Kheradpisheh, M.~Ghodrati, M.~Ganjtabesh, and T.~Masquelier.
\newblock Deep networks can resemble human feed-forward vision in invariant
  object recognition.
\newblock {\em Scientific Reports}, 6:32672, Sept. 2016.

\bibitem{Kim2014-fp}
J.~S. Kim, M.~J. Greene, A.~Zlateski, K.~Lee, M.~Richardson, S.~C. Turaga,
  M.~Purcaro, M.~Balkam, A.~Robinson, B.~F. Behabadi, M.~Campos, W.~Denk, H.~S.
  Seung, and {EyeWirers}.
\newblock Space-time wiring specificity supports direction selectivity in the
  retina.
\newblock {\em Nature}, 509(7500):331--336, 15~May 2014.

\bibitem{Kovashka2016-oi}
A.~Kovashka, O.~Russakovsky, L.~Fei-Fei, and K.~Grauman.
\newblock Crowdsourcing in computer vision.
\newblock {\em Foundations and Trends in Computer Graphics and Vision},
  10(3):177--243, 2016.

\bibitem{Kummerer2016-vp}
M.~K{\"{u}}mmerer, T.~S.~A. Wallis, and M.~Bethge.
\newblock {DeepGaze} {II}: Reading fixations from deep features trained on
  object recognition.
\newblock {\em arXiv}, 1610.01563, 5~Oct. 2016.

\bibitem{Lake2015-jw}
B.~M. Lake, W.~Zaremba, R.~Fergus, and T.~M. Gureckis.
\newblock Deep neural networks predict category typicality ratings for images.
\newblock In {\em Proceedings of the 37th Annual Conference of the Cognitive
  Science Society}, 2015.

\bibitem{Lin_undated-qc}
T.-Y. Lin, M.~Maire, S.~Belongie, J.~Hays, P.~Perona, D.~Ramanan,
  P.~Doll\'{a}r, and C.~Lawrence~Zitnick.
\newblock Microsoft {COCO}: Common objects in context.
\newblock In {\em European Conference on Computer Vision ({ECCV})}, pages
  740--755. Springer International Publishing, 2014.

\bibitem{Murray2011-hl}
R.~F. Murray.
\newblock Classification images: A review.
\newblock {\em J. Vis.}, 11(5):2--, 1~Jan. 2011.

\bibitem{Nielsen2008-ey}
K.~J. Nielsen, N.~K. Logothetis, and G.~Rainer.
\newblock Object features used by humans and monkeys to identify rotated
  shapes.
\newblock {\em J. Vis.}, 8(2):9.1--15, 22~Feb. 2008.

\bibitem{Papadopoulos_undated-tp}
D.~P. Papadopoulos, A.~D.~F. Clarke, F.~Keller, and V.~Ferrari.
\newblock Training object class detectors from eye tracking data.
\newblock In {\em European Conference on Computer Vision ({ECCV})}, pages
  361--376. Springer International Publishing, 2014.

\bibitem{Peterson2016-ig}
J.~C. Peterson, J.~T. Abbott, and T.~L. Griffiths.
\newblock Adapting deep network features to capture psychological
  representations.
\newblock {\em arXiv}, 1608.02164, 6~Aug. 2016.

\bibitem{Russakovsky2014-ff}
O.~Russakovsky, J.~Deng, H.~Su, J.~Krause, S.~Satheesh, S.~Ma, Z.~Huang,
  A.~Karpathy, A.~Khosla, M.~Bernstein, A.~C. Berg, and L.~Fei-Fei.
\newblock {ImageNet} large scale visual recognition challenge.
\newblock {\em CoRR}, abs/1409.0:43, 1~Sept. 2014.

\bibitem{Saleh2016-hs}
B.~Saleh, A.~Elgammal, and J.~Feldman.
\newblock The role of typicality in object classification: Improving the
  generalization capacity of convolutional neural networks.
\newblock {\em arXiv}, 1602.02865, 9~Feb. 2016.

\bibitem{Scheirer2014-gp}
W.~J. Scheirer, S.~E. Anthony, K.~Nakayama, and D.~D. Cox.
\newblock Perceptual annotation: Measuring human vision to improve computer
  vision.
\newblock {\em IEEE Trans. Pattern Anal. Mach. Intell.}, 36(8):1679--1686, Aug.
  2014.

\bibitem{Schyns2002-xt}
P.~G. Schyns, L.~Bonnar, and F.~Gosselin.
\newblock Show me the features! {U}nderstanding recognition from the use of
  visual information.
\newblock {\em Psychol. Sci.}, 13(5):402--409, Sept. 2002.

\bibitem{Shanmuga_Vadivel2015-wi}
K.~Shanmuga~Vadivel, T.~Ngo, M.~Eckstein, and B.~S. Manjunath.
\newblock Eye tracking assisted extraction of attentionally important objects
  from videos.
\newblock In {\em Proceedings of the {IEEE} Conference on Computer Vision and
  Pattern Recognition}, pages 3241--3250, 2015.

\bibitem{Simonyan2014-ap}
K.~Simonyan, A.~Vedaldi, and A.~Zisserman.
\newblock Deep inside convolutional networks: Visualising image classification
  models and saliency maps.
\newblock In {\em {ICLR} Workshop}, 2014.

\bibitem{Simonyan2015-pj}
K.~Simonyan and A.~Zisserman.
\newblock Very deep convolutional networks for {Large-Scale} image
  recoginition.
\newblock {\em Intl. Conf. on Learning Representations (ICLR)}, pages 1--14,
  2015.

\bibitem{Steggink2010-lb}
J.~Steggink and C.~G.~M. Snoek.
\newblock Adding semantics to image-region annotations with the {Name-It-Game}.
\newblock {\em Multimedia Systems}, 17(5):367--378, 9~Dec. 2010.

\bibitem{Ullman2016-ea}
S.~Ullman, L.~Assif, E.~Fetaya, and D.~Harari.
\newblock Atoms of recognition in human and computer vision.
\newblock {\em Proceedings of the National Academy of Sciences of the USA},
  13(10):2744–2749, 2016.

\bibitem{Vig_undated-jg}
E.~Vig, M.~Dorr, and D.~Cox.
\newblock Space-variant descriptor sampling for action recognition based on
  saliency and eye movements.
\newblock In {\em European Conference on Computer Vision ({ECCV})}, pages
  84--97, 2012.

\bibitem{Vijayanarasimhan2014-em}
S.~Vijayanarasimhan and K.~Grauman.
\newblock {Large-Scale} live active learning: Training object detectors with
  crawled data and crowds.
\newblock {\em Int. J. Comput. Vis.}, 108(1-2):97--114, 1~May 2014.

\bibitem{Von_Ahn2004-qe}
L.~von Ahn and L.~Dabbish.
\newblock Labeling images with a computer game.
\newblock In {\em Proceedings of the {SIGCHI} Conference on Human Factors in
  Computing Systems}, CHI '04, pages 319--326, New York, NY, USA, 2004. ACM.

\bibitem{Von_Ahn2006-bz}
L.~von Ahn, R.~Liu, and M.~Blum.
\newblock Peekaboom: A game for locating objects in images.
\newblock In {\em Proceedings of the {SIGCHI} Conference on Human Factors in
  Computing Systems}, CHI '06, pages 55--64, New York, NY, USA, 2006. ACM.

\bibitem{Vondrick2015-pr}
C.~Vondrick, H.~Pirsiavash, A.~Oliva, and A.~Torralba.
\newblock Learning visual biases from human imagination.
\newblock In C.~Cortes, N.~D. Lawrence, D.~D. Lee, M.~Sugiyama, and R.~Garnett,
  editors, {\em Advances in Neural Information Processing Systems 28}, pages
  289--297. Curran Associates, Inc., 2015.

\bibitem{Yamins2016-yx}
D.~L.~K. Yamins and J.~J. DiCarlo.
\newblock Using goal-driven deep learning models to understand sensory cortex.
\newblock {\em Nat. Neurosci.}, 19(3):356--365, 23~Feb. 2016.

\bibitem{Yarbus1967-xp}
A.~L. Yarbus.
\newblock {\em Eye Movements and Vision}.
\newblock Plenum press, New York, 1967.

\bibitem{Zeiler2013-mk}
M.~D. Zeiler and R.~Fergus.
\newblock Visualizing and understanding convolutional networks.
\newblock {\em arXiv preprint arXiv:1311.2901}, pages 818--833, 2013.

\bibitem{zhou2015cnnlocalization}
B.~Zhou, A.~Khosla, L.~A., A.~Oliva, and A.~Torralba.
\newblock {Learning Deep Features for Discriminative Localization.}
\newblock {\em CVPR}, 2016.

\end{thebibliography}
}

\end{document}